\documentclass{ecai}  
\usepackage{tikz}
\usetikzlibrary{positioning, 
                quotes}
\usepackage{graphicx}
\usepackage{latexsym}
\usepackage{amsmath}

\usepackage{breqn}
\usepackage{rotating}
\usepackage{verbatim}
\usepackage{multirow}

\newcommand{\true}{\texttt{tt}}
\newcommand{\false}{\texttt{ff}}

\newcommand{\moddiam}[1]{\langle#1\rangle}
\newcommand{\prefr}{\prec}
\newcommand{\prefl}{\succ}

\newcommand{\mlangpref}[3]{#1\ \texttt{P}_{#3}\ #2}

\newcommand{\lsem}[2]{\langle\![#1]\!\rangle_{#2}}
\newcommand{\msem}[2]{[\!\![#1]\!\!]_{#2}}

\newcommand{\trans}[1]{\stackrel{#1}{\rightarrow}}

\newcommand{\ipg}{I}

\newcommand{\collaba}{Collective} 
\newcommand{\collabb}{Constructive}

\newtheorem{example}{Example}
\newtheorem{definition}{Definition}

\begin{document}

\begin{frontmatter}

\title{Representing and Reasoning with Multi-Stakeholder Qualitative Preference Queries}

\author[A]{\fnms{Samik}~\snm{Basu}
\thanks{Corresponding Author. Email: sbasu@cs.iastate.edu.}}
\author[B]{\fnms{Vasant}~\snm{Honavar}\orcid{0000-0001-5399-3489}} 
\author[C]{\fnms{Ganesh Ram}~\snm{Santhanam}}
\author[D]{\fnms{Jia}~\snm{Tao}\orcid{0000-0002-2342-3271}}

\address[A]{Department of Computer Science, Iowa State University, Ames, IA, USA}
\address[B]{College of Information Sciences and Technology, Pennsylvania State University, University Park, PA, USA}
\address[C]{Department of Electrical and Computer Engineering, Iowa State University, Ames, IA, USA}
\address[D]{Department of Computer Science, Lafayette College, Easton, PA, USA}

\begin{abstract}
Many decision-making scenarios, e.g., public policy, healthcare, business, and disaster response, require accommodating the preferences of multiple stakeholders. We offer the first formal treatment of reasoning with multi-stakeholder qualitative preferences in a setting where stakeholders express their preferences in a qualitative preference language, e.g., CP-net, CI-net, TCP-net, CP-Theory. We introduce a query language for expressing queries against such preferences over sets of outcomes that satisfy  specified criteria, e.g., $\mlangpref{\psi_1}{\psi_2}{A}$ (read loosely as the set of outcomes satisfying $\psi_1$ that are preferred over outcomes satisfying $\psi_2$ by a set of stakeholders $A$).  
Motivated by practical application scenarios, we introduce and analyze several alternative semantics for such queries, and examine their interrelationships.  We provide a provably correct algorithm for answering multi-stakeholder qualitative preference queries using model checking in alternation-free $\mu$-calculus. We present experimental results that demonstrate the feasibility of our approach.
\end{abstract}

\end{frontmatter}

\section{Introduction}
\label{sec:intro}

The ability to express and reason about preferences over a set of alternatives is central to rational decision-making in a broad range of applications, including software design~\cite{vanLamsweerde:GORE,Liaskos:RE10,Sohrabi:AAAI2011,Sohrabi:ISWC10,fattah2021cpnet,abdulaziz:21},  public policy, e.g., city planning~\cite{CityPlanning:2016,DBLP:conf/birthday/SonPB14}, healthcare \cite{HealthCare:1998},  security \cite{Bistarelli:07,Gunasekharan:CISRC17}, privacy \cite{Oster:FACS2012}, among others.  In general, the preferences can be quantitative~\cite{Keeney:Camb97,French:EH86} or qualitative~\cite{Brafman:AIMag09,Doyle:AIMag99}. But 
stakeholders often find it natural to
express their preferences in qualitative terms \cite{Santhanam:book2016}, e.g., that a cheaper car is preferred to a more expensive car. Hence, there has been a growing interest in languages and tools for representing and reasoning with qualitative preferences\cite{Domshlak:AI2011,Santhanam:book2016,pcpnet:Cornelio:JAIR21}. 
For example, ~\cite{Santhanam:AAAI2010} leverage advances in model checking ~\cite{Clarke:MIT2000,Queille:1982,Cimatti:CAV2002} to provide efficient and hence practically useful tools for reasoning with the qualitative preferences of single stakeholders \cite{CRISNER,Santhanam:Arxiv2015}. 

However, decision-making in real-world settings often needs to accommodate the preferences of \textit{multiple} stakeholders. Consider, for example, the task of choosing a care plan for a critically ill patient. The stakeholders, in this case, may include the patient concerned with their health outcome and the cost of care, the physician committed to ensuring that the patient receives the best care available, the family members with an interest in the patient's well-being, the hospital system seeking to maximize its profits, and the insurance provider seeking to minimize the reimbursements. 
A key challenge in extending the preference representation languages and reasoning tools from the single stakeholder setting to the multi-stakeholder setting has to do with maintaining, and reasoning with the (possibly conflicting) preferences of stakeholders. Furthermore, the preferences of some stakeholders in some settings may override those of others, e.g., due to their relative roles in an organization, or due to differences in their expertise as it relates to specific aspects of the application domain, etc. Ensuring transparency and accountability of decision-making requires that the system be able to explain {\em how} the stakeholders' preferences impact the outcomes. 



\smallskip
\noindent\textbf{Contributions.\ } The key contributions of the paper are as follows:
(i) We provide the first formal treatment of reasoning with multi-stakeholder qualitative preferences. We consider the setting where the stakeholders express their preferences in a qualitative preference language, e.g., CP-net, CI-net, TCP-net, CP-Theory.
(ii) We introduce a query language for expressing queries with respect to the preferences of multiple stakeholders over outcomes that satisfy a set of specified criteria. 
(iii) We generalize the {\em induced preference graphs} that encode the qualitative preferences of a single stakeholder to {\em multi-stakeholder induced preference graphs} that encode the preferences of multiple stakeholders.  
(iv) We introduce and analyze several alternative semantics for such queries, motivated by the needs of different application scenarios, and examine their inter-relationships.  
(v)
We provide a provably correct algorithm for answering multi-stakeholder preference queries using model checking in $\mu$-calculus; and
(vi) We present results of experiments that demonstrate the feasibility of our approach.

\section{Qualitative Preference Languages}
\label{sec:preflanguage}
We consider settings in which stakeholders express preferences over a set of alternatives or outcomes, where each alternative is described by a set of attributes or (preference) variables. Stakeholders may directly express their preference between a pair of alternatives, by asserting that one valuation of the variables is preferred to another. In addition, preferences over sets of alternatives may be succinctly stated over (a) the possible valuations of each variable, i.e., \textit{intra-variable preference}; or (b) the variables themselves indicating their \textit{relative importance}.  Several qualitative preference languages with varying expressive power have been studied in the literature. For instance, CP-nets \cite{Boutilier:JAIR04} allow the expression of preferences over the valuations of each variable as a strict partial order, possibly conditioned on specific valuation(s) of one or more other variables. TCP-nets \cite{Brafman:JAIR06} extend CP-nets by additionally allowing expression of the relative importance of one variable over another. CP-theories \cite{Wilson:AAAI2004} further extend TCP-nets by allowing the expression of the relative importance of one variable over a set of variables. 

Formally, let $X = \{X_i \,|\, 0 < i \leq n\}$ be a set of preference variables, $D_i$ be the domain of $X_i$, and $v_i$ be the assignment of $X_i$ to a particular valuation in $D_i$. Let $O = \Pi_{X_i\in X} D_i$ be the set of alternatives or outcomes, and $O^P = \Pi_{X_i \in Y \subseteq X} D_i$ be the set of partial alternatives or outcomes. Each outcome $o \in O$ is represented as a tuple of valuations of each variable, i.e., $o = \langle v_1, v_2, \dots v_n \rangle$. We use the following notation to represent a preference statement 
\[
P:\ [c]\ (X_i = v_i) \succ (X_i = {v_i}')\  [Y]
\] 
where $c \in O^P$ is the condition under which this preference over $X_i$'s valuation holds, and $Y \subseteq X\setminus {X_i}$ is the set of variables less important than $X_i$. 
For brevity, we drop $[c]$ when $c = true$ and $[Y]$ when $Y = \emptyset$. A preference statement $P$ specifies that when $c$ holds, the valuation $v_i$ is preferred to ${v_i}'$ for variable $X_i$, regardless of the valuations and intra-variable preferences of the variables in $Y$.

\begin{example}
\label{ex:running}
Consider the preferences of a set of stakeholders  tasked with prioritizing vulnerabilities to be mitigated as part of protecting a critical network. Each vulnerability may be described by three variables describing the threats it poses, namely (a) attack complexity (A) with values Simple or Complex (indicating whether the complexity of the attack required to exploit the vulnerability is low or high); (b) exploit availability (E) with values Code or No-Code (indicating whether code to exploit the vulnerability is available); and (c) fix availability (F) for the vulnerability with values Fix or No-Fix (indicating whether a fix can be applied or not). Figure \ref{fig:pref-statements} shows some preferences with respect to these variables. Note that $P_5$ is a direct preference between two alternatives, $P_7$ is a relative importance preference, and the rest specify intra-variable preferences. Now consider three stakeholders, say, $1$, $2$, and $3$. Suppose stakeholder $1$ holds the preferences $P_1$ and $P_2$ of the \textit{incident-response} team whose overall goal is to prioritize readily exploitable vulnerabilities with no available fixes when initiating an immediate response, e.g., disconnecting critical systems from the network. Suppose stakeholder $2$ holds the preferences $P_3$, $P_4$ and $P_5$ of the \textit{patch-adaptation} team responsible for adapting existing fixes to address the vulnerability (hence has preferences conditioned on the fixed availability). Finally, suppose stakeholder $3$ holds the preferences $P_1$, $P_6$ and $P_7$ of the \emph{severity-assessment} team that aims to prioritize exploitable vulnerabilities based on their severity for action by the incident-response team.
\end{example}

\begin{figure}
    \centering
\begin{center}
  \begin{tabular}{ll}
    $P_1$ & $E=Code \succ_E E=No-Code$ \\
    $P_2$ & $[E=Code] \ \  F=No-Fix \succ_F F=Fix$ \\
    $P_3$ & $[F=Fix] \ \  E=Code \succ_E E=No-Code$ \\
    $P_4$ & $[F=Fix] \ \  A=Simple \succ_A A=Complex$ \\
    $P_5$ & $\langle E=No-Code, A=Simple, F=No-Fix\rangle  \ \ \succ $ \\ 
    $ $ &  $\ \ \   \ \ \ \langle E=Code, A=Complex, F=No-Fix\rangle$ \\
    $P_6$ & $A=Simple \succ_A A=Complex$ \\
    $P_7$ & $E=Code \succ_E E=No-Code [A, F]$ \\
  \end{tabular}
\end{center}
    \caption{Preference statements}
    \label{fig:pref-statements}
\end{figure}

\begin{figure}
  \scalebox{0.25}
  {\includegraphics{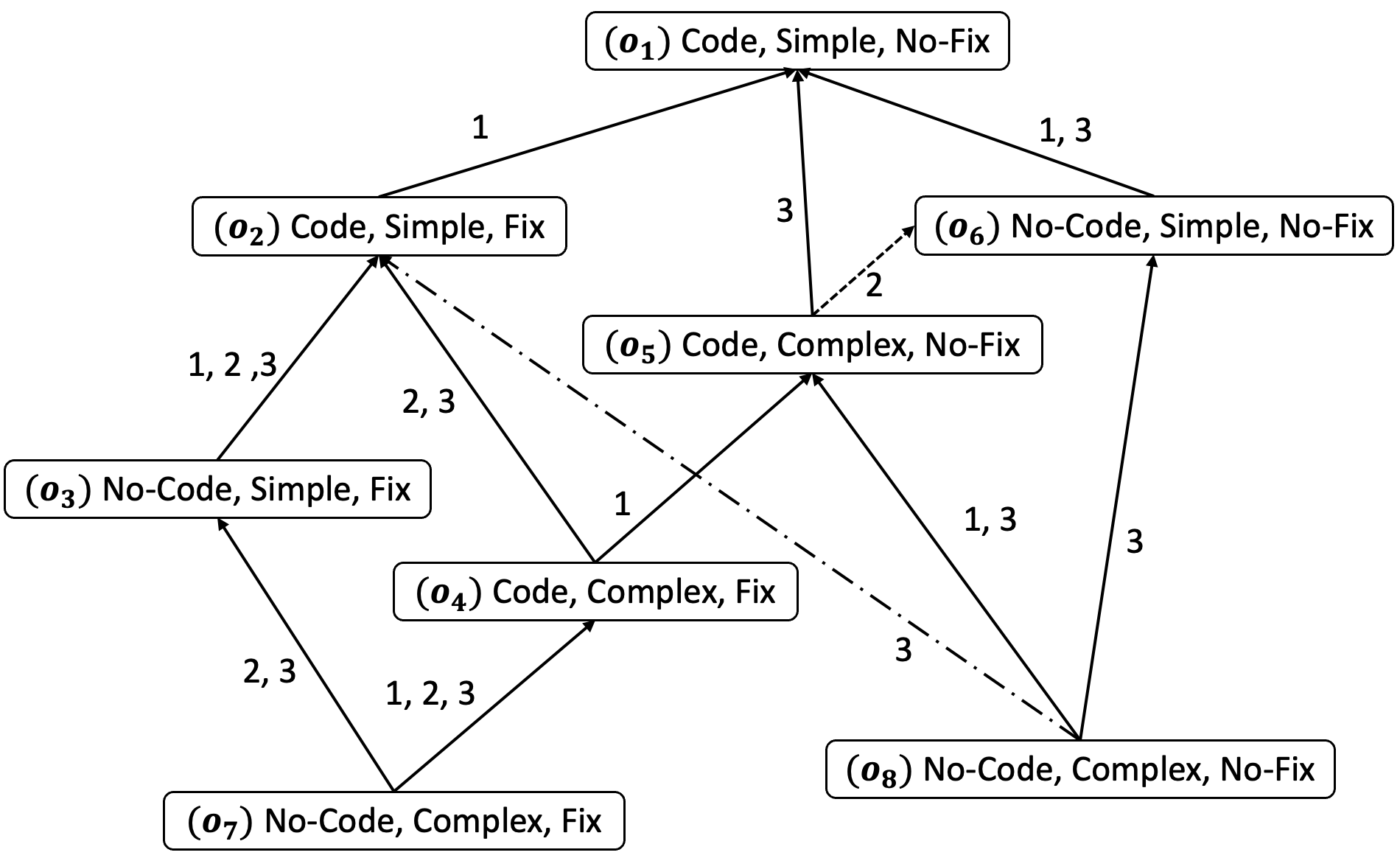}}
\caption{Induced preference graph}
\label{fig:induced-pref-graph}
\end{figure}

\noindent\textbf{Semantics of Preferences. }
\label{sec:ipg}
The semantics of CP-nets, TCP-nets, and CP-theories is based on and extends the \textit{ceteris-paribus} principle \cite{Hansson:Springer1995}. The preference statements induce a strict partial order over the alternatives. For instance, for $o, o' \in O$, a preference statement $P$: $[c]$ $(X_i=v_i) \succ (X_i={v_i}')$ induces a preference from $o'$ to $o$ (denoted $o' \prefr o$) if both satisfy $c$; their valuations for $X_i$ are $v_i$ and ${v_i}'$ respectively; and their valuations for all other variables are identical.

\begin{definition}[Induced Preference Graph]
  Given a set of outcomes $O$ described by a set $AP$ of propositional variables, an induced preference graph $\ipg = (O\ \cup\ \{\bot\}, E, L)$ is defined over $O\cup\{\bot\}$ with an edge relation $E\subseteq (O\cup\{\bot\}) \times (O\cup\{\bot\})$ and a labeling function that maps each element in $O\cup\{\bot\}$ to a subset of propositional variables $L:(O\cup\{\bot\})\rightarrow \mathcal{P}(AP)$.  An edge
  $e=(o_1, o_2) \in E$ captures the fact that $o_1\prefr o_2$ and
  there exists a flip in the valuation of exactly one variable that
  contributes to this preference. For each $o\in O$, there exists 
  an edge ($\bot$, o), indicating that every outcome is preferred to $\bot$.
  Furthermore, $L(\bot) = \emptyset$ indicates that the $\bot$ does not satisfy
  any atomic proposition. 
  \label{def:ipg}
\end{definition}

\begin{definition}[Multi-Stakeholder Induced Preference Graph] A multi-stakeholder
  induced preference graph is an induced preference graph where each
  edge in the graph is annotated by the set of stakeholders whose preferences
  induce that edge. That is, $\ipg = (O\cup\{\bot\}, E, L, \mathcal{A})$ where the edge
  relation $E \subseteq (O\cup\{\bot\})\times \mathcal{P}(\mathcal{A}) \times (O\cup\{\bot\})$. An edge
  $e=(o_1, A, o_2) \in E$ captures the fact that $o_1\prefr o_2$ for
  every agent in $A$. We note that $e=(\bot, \mathcal{A}, o)$ for every $o\in O$.
  \end{definition}
  
\begin{example}
  \label{ex:ipg}
  The  (partial view of) induced preference graph of the preferences stated in Figure~\ref{fig:pref-statements} is given in Figure~\ref{fig:induced-pref-graph}. The edges correspond to flips from the less preferred to the more preferred alternative and are labeled with the preferences induced by the corresponding stakeholders. For instance, the edge from $o_4$ to $o_5$ is induced by the preference statement $P_2$ of stakeholder $1$. Similarly, the edge from $o_5$ to $o_6$ is induced by $P_5$ of stakeholder $2$ and the edge from $o_8$ to $o_2$ is induced by $P_7$ of stakeholder $3$. Note that some edges induced by stakeholder $3$'s preferences and the edges from $\bot$ to all of the outcomes are omitted for the sake of readability.
\end{example}
  
We will denote the edges in $\ipg$ as $o_1\trans{A} o_2$, where
$A$ is the set of agents whose preferences have induced the edge from
$o_1$ to $o_2$. 

\begin{definition}[$\prefr_A$ and $\prefr_A^+$]\label{def:path}
We write $o\prefr_A o'$ if there exists an edge
$o\trans{A'}o'$ and $A\cap A' \neq \emptyset$. Similarly, 
$o\prefr_A^+ o'$ if there exists a path
$o=o_1\trans{A_1}o_2\trans{A_2}\ldots \trans{A_k}o_{k+1}=o'$
where $\forall i\in [1,k]. (A\cap A_i)\neq \emptyset$.

\end{definition}
When $A$ is singleton ($A= \{a\}$), we will write
$o\prefr_a o'$.

\section{Single Stakeholder Preference Queries}
\label{sec:lang}

We first introduce a language for expressing queries with respect to single stakeholder qualitative preferences  before proceeding to consider multi-stakeholder preferences queries. A key feature of this language is that it allows expressing queries against preferences over properties of outcomes, rather than the outcomes themselves. Thus, it can readily accommodate preferences expressed in existing  qualitative preference languages such as CP-nets \cite{Boutilier:JAIR04}, TCP-nets \cite{Brafman:JAIR06}, and CP-theories \cite{Wilson:AAAI2004}. This allows us, for example, to query for outcomes with properties that are more preferred to all other outcomes. The resulting single stakeholder preference query language can express a range of preference queries (e.g., find the set of non-dominated outcomes) of common 
interest.\\

\noindent\textbf{Syntax. }
The syntax of the query language is described over atomic
propositions, propositional constants, boolean connectives
and a (new) operator \texttt{P}: preference operator over properties. 
The language $\Psi$ is defined by the grammar: 
\[
\psi \rightarrow \true~|~\false~|~\texttt{AP}~|~\neg\psi~|~\psi\land\psi~|~\psi\lor\psi~|~\mlangpref{\psi}{\psi}{a}
\]
The answer to a query corresponds to the set of outcomes that belongs to the semantics of the query. 
For instance, all outcomes are returned for a query $\true$, while no outcome is returned for the query $\false$. A query involving an atomic proposition simply returns the outcomes that satisfy the proposition. Answers to queries involving Boolean connectives conform to the natural meaning of the connectives. The query $\mlangpref{\psi_1}{\psi_2}{a}$ returns the outcomes satisfying $\psi_1$ that are more preferred than outcomes satisfying $\psi_2$ based on the preferences of the stakeholder $a$.\\

\noindent\textbf{Semantics. }
The semantics of the query language is defined over the set of outcomes (states) in the preference graph $\ipg$ induced by the given preferences. Let $\mathcal{I}$ be the set of all preference graphs that can be induced by single stakeholder preferences with respect to which single stakeholder queries can be expressed given the syntax described above.
We use the (semantic) function $\lsem{}{}:\Psi \times\ \mathcal{\ipg} \rightarrow
\mathcal{P}(O)$, to define the semantics of $\psi \in \Psi$ in the
context of an induced preference graph $\ipg \in \mathcal{\ipg}$.
That is, $\lsem{\psi}{\ipg}$ denotes the set of outcomes in
$\ipg$ that satisfy the query expressed using the formula $\psi$.
We will omit $\ipg$ from the definition unless it is explicitly
necessary to distinguish between semantics in the context of two different induced preference graphs.
\[
\begin{array}{rcl}
\lsem{\true}{} & = & O \\

\lsem{\false}{} & = & \emptyset \\

\lsem{p}{} & = & \{o~|~p \in L(o)\} \\

\lsem{\neg\psi}{} & = & O - \lsem{\psi}{} \\

\lsem{\psi_1\land\psi_2}{}  & = & \lsem{\psi_1}{} \ \cap\ \lsem{\psi_2}{} \\

\lsem{\mlangpref{\psi_1}{\psi_2}{a}}{} & = & 
\lsem{\psi_1}{}\ \cap \
\{o~|~\exists o'.o' \in \lsem{\psi_2}\ \land\ o \prefl^+_a o'\} \\
& & \ \ \ \ \ \ \ \ \cap \
 \{o~|~\forall o'.o' \in \lsem{\psi_2}\ \Rightarrow\ o \not\prefr^+_a o'\} \\
\end{array}
\]
Propositional constants $\true$ and $\false$ are satisfied by all and
no outcomes, respectively. The proposition $p$ is satisfied by any
outcome that satisfies $p$. The formulas over Boolean connectives
(negation, conjunction, disjunction) conform to the standard set-based
semantics (complement, intersection, union). The formula
$\mlangpref{\psi_1}{\psi_2}{a}$ is satisfied by outcomes that (i)
satisfy $\psi_1$, (ii) are preferred to at least one outcome that
satisfies $\psi_2$, and (iii) are not less preferred to any outcome that
satisfies $\psi_2$ by the stakeholder $a$. In short,  $\mlangpref{\psi_1}{\psi_2}{a}$ is the set of outcomes satisfying $\psi_1$ that are more preferred to
outcomes satisfying $\psi_2$ by the stakeholder $a$.\\

\noindent The resulting query language can be used to express queries such as:

\begin{itemize}
\item
  What is the set of outcomes that are preferred  by the stakeholder $a$ to outcomes that satisfy $\psi$? The is expressed as $\mlangpref{\true}{\psi}{a}$.
  
\item What is the non-dominated set of outcomes relative to stakeholder $a$'s preferences? The query can be expressed as $\mlangpref{\true}{\true}{a}$. What is the non-dominated set of outcomes for stakeholder $a$ that satisfies $\psi$? This can be  expressed as $\mlangpref{\psi}{\true}{a}$.

\item With respect to stakeholder $a$'s preferences, what are the best improvements to outcomes satisfying $\psi$?  The query can be expressed as $(\mlangpref{\true}{\true}{a})\ \land\ (\mlangpref{\true}{\psi}{a})$.
  \end{itemize}

\begin{example}
  If $\psi = \texttt{Code}$, then the semantics
  of $\mlangpref{\true}{\psi}{1}$  (for stakeholder $1$)
  is the set of outcomes $\{o_1, o_5\}$. This is because, 
  while both $o_1$ and $o_5$ dominate some outcome satisfying
  $\texttt{Code}$  with respect to stakeholder $1$'s preferences, they are not dominated by any outcome that satisfies \texttt{Code}. On the other hand,
  the query $\mlangpref{\true}{\psi}{2}$  (for stakeholder $2$)
  yields the set $\{o_2, o_6\}$.
  \label{ex:basicquery}
\end{example}

\begin{example}
For stakeholder $1$, the non-dominated
set of outcomes is $\{o_1, o_5\}$ (result of the query: $\mlangpref{\true}{\true}{1}$), while for stakeholder $2$, the non-dominated set is $\{o_1, o_2, o_6, o_8\}$.
Note that the outcome $o_8$ neither dominates nor is  dominated by any outcome, However, it dominates $\bot$ and hence is included as part of the non-dominated set. 
  \label{ex:singlenondom}
\end{example}
\noindent\textbf{Cycles in Induced Preference Graphs.\ }
Cycles in an induced preference graph are indicative of inconsistencies
in the underlying preferences, the result being some outcome $o$ both more and less preferred to an outcome $o'$.
Does this pose any inconsistencies in the semantic interpretation of
$\mlangpref{\psi_1}{\psi_2}{a}$, when $o$ satisfies $\psi_1$ and $o'$ satisfies $\psi_2$?
The answer is no. This is because semantics of $\mlangpref{\psi_1}{\psi_2}{a}$ excludes all outcomes that are less preferred to outcomes satisfying $\psi_2$. Hence, the outcome $o$ will not be included in the set of outcomes returned by $\mlangpref{\psi_1}{\psi_2}{a}$ as it is less preferred to $o'$.

\section{Multi-Stakeholder Preference Queries}
\label{sec:multis}

We proceed to extend the preceding language for expressing preference queries  to allow preference queries with respect to the preferences of a set of stakeholders, as opposed to just a single stakeholder. Specifically, we add a new query construct $\mlangpref{\psi_1}{\psi_2}{A}$ where $A \subseteq \mathcal{A}$, where $\mathcal{A}$ is
the set of all stakeholders. When $A$ is a singleton $a$, we
use $\mlangpref{\psi_1}{\psi_2}{a}$ to denote the query about the preferences of a single stakeholder $a$ (as described in Section~\ref{sec:lang}).  In what follows, we  describe the semantics of multi-stakeholder preference queries  under several alternative interpretations of multi-stakeholder preferences.\\

\noindent\textbf{Consensus Semantics.}
Consensus semantics, as the name suggests, is defined as the set of
outcomes, whose preference over another set of outcomes, is decided by agreement among the set of stakeholders in question. Formally,
\[
\lsem{\mlangpref{\psi_1}{\psi_2}{A}}{}^{cs} = \displaystyle\bigcap_{a\in A} \lsem{\mlangpref{\psi_1}{\psi_2}{a}}{}
\]
\begin{example}
In Example~\ref{ex:basicquery}, as per the consensus semantics
the result of the query $(\mlangpref{\true}{\texttt{Code}}{\{1,2\}})$ is the
empty set as the stakeholders $1$ and $2$ do not agree on the outcomes
that are more desirable than outcomes satisfying \texttt{Code}. On the other hand,
stakeholders $1$ and $2$ agree on the non-dominated set $\{o_1\}$ computed
as the semantics of $\mlangpref{\true}{\true}{\{1,2\}}$ (see Example
\ref{ex:singlenondom}).

\label{ex:consensus}
\end{example}

\noindent\textbf{Collaborative Semantics. }
Unlike consensus semantics, which requires a complete agreement
among the stakeholders, a collaborative semantics allows
the stakeholders to arrive at a compromise that is not disagreeable to any stakeholder. There are several ways to realize such a compromise that correspond to different interpretations of the semantics of $\mlangpref{\psi_1}{\psi_2}{A}$. Recall that   $\mlangpref{\psi_1}{\psi_2}{A}$ must return the set of outcomes that (i) satisfy $\psi_1$, (ii) are preferred to at least one outcome that satisfies $\psi_2$, and (iii) are not less preferred to any outcome that satisfies $\psi_2$.
We will refer to the last two conditions (ii and iii) as follows:
\begin{enumerate}
\item \emph{Witness Condition ($\mathtt{W}$)\ }  for determining the set of outcomes that are preferred to at least one outcome satisfying
$\psi_2$.
\item \emph{Agreement Condition ($\mathtt{A}$)\ } for determining
the set of outcomes that are not less preferred to any outcome satisfying
$\psi_2$.
\end{enumerate}
Each of these conditions can be collaboratively decided in two ways: 
 
\begin{enumerate}
\item \emph{\collaba\ Collaboration.} 
The set of outcomes that are preferred to at least one outcome satisfying $\psi$ is chosen to be the \textbf{union} of outcomes preferred by each of the stakeholders
to {\em at least one outcome} satisfying $\psi$. 

\item \emph{\collabb\ Collaboration.\ }  
An outcome $o'$ is considered to be preferred to outcome $o$ when there exists a path
in the induced preference graph from $o$ to $o'$ where each edge along the path may be induced by the preferences of one or more stakeholders. Thus, there is no requirement that all of the edges along the path be induced by the preferences of the same stakeholder. Hence, the stakeholders collaboratively construct the path from $o$ to $o'$ by contributing one or more edges to the path based on their individual preferences. This can be viewed as \textbf{chaining}
induced preference edges of different
stakeholders to arrive at the result.
\end{enumerate}
\collabb\ Collaboration is useful in situations
where each stakeholder may not have complete information or expertise to determine a dominance relation between a pair of outcomes but they may be able to collaborate to arrive at a conclusion. For instance, healthcare providers (doctors, nurses) and hospital administrators may collaborate to develop an optimal placement strategy for hand sanitizers in the hospital. The healthcare providers present their preferences based on their knowledge of the usage of hand sanitizers at different times and locations, whereas the hospital administrators present their preferences based on the cost of procuring hand sanitizers. 

Now we have two different choices for the witness ($\mathtt{W}$) condition and agreement ($\mathtt{A}$) condition: 
\begin{itemize}
    \item[$\mathtt{W}_1$.\ ] \collaba\ collaboration for deciding witness condition for $\psi_2$
    in $\mlangpref{\psi_1}{\psi_2}{A}$:
    \[
    \displaystyle\bigcup_{a\in A}\{o~|~\exists o'.o' \in \lsem{\psi_2}\ \land\ o \prefl_a^+ o'\} 
    \]
    \item[$\mathtt{W}_2$.\ ] \collabb\ collaboration for deciding witness condition for $\psi_2$
    in $\mlangpref{\psi_1}{\psi_2}{A}$:
    \[
    \{o~|~\exists o'.o' \in \lsem{\psi_2}\ \land\ o \prefl_A^+ o'\} 
    \]
    \item[$\mathtt{A}_1$.\ ] \collaba\ collaboration for deciding agreement condition for $\psi_2$
    in $\mlangpref{\psi_1}{\psi_2}{A}$:
    \[
    \begin{array}{ll}
    & O\setminus \displaystyle\bigcup_{a\in A}\{o~|~\exists o'.o' \in \lsem{\psi_2}\ \land\ o \prefr_a^+ o'\} \\[1.25em]
    = & \displaystyle\bigcap_{a\in A}\{o~|~\forall o'.o' \in \lsem{\psi_2}\ \Rightarrow\ o \not\prefr_a^+ o'\} 
    \end{array}
    \]
    
    \item[$\mathtt{A}_2$.\ ] \collabb\ collaboration for deciding agreement condition for $\psi_2$:
    in $\mlangpref{\psi_1}{\psi_2}{A}$
    \[
    \begin{array}{ll}
    & O\setminus \{o~|~\exists o'.o' \in \lsem{\psi_2}\ \land\ o \prefr_A^+ o'\}  \\[0.5em]
    = &
    \{o~|~\forall o'.o' \in \lsem{\psi_2}\ \Rightarrow\ o \not\prefr_A^+ o'\}
    \end{array}
    \]
\end{itemize}

\begin{example}
Consider the induced preference graph in Figure~\ref{fig:induced-pref-graph}.
For stakeholder $1$, the set of outcomes that dominate the outcomes
satisfying \texttt{No-Code} is $\{o_1, o_2, o_4, o_5\}$. This is because
$o_4\prefl_1 o_7$, $o_5\prefl_1 o_4$, $o_2\prefl_1 o_3$ and $o_1\prefl_1 o_2$.
On the other hand, for stakeholder $2$, the set of outcomes that
dominate the outcomes satisfying \texttt{No-Code} is $\{o_2, o_3, o_4\}$. 

Therefore, for $\psi_2 = \texttt{No-Code}$,  we have:
\[
\mathtt{W}_1: \displaystyle\bigcup_{a\in \{1,2\}}\{o~|~\exists o'.o' \in \lsem{\psi_2}\ \land\ o \prefl_a^+ o'\}  = \{o_1, o_2, o_3, o_4, o_5\}
\]
On the other hand, 
\[
\mathtt{W}_2: \{o~|~\exists o'.o' \in \lsem{\psi_2}\ \land\ o \prefl_{\{1,2\}}^+ o'\} = \{o_1, o_2, o_3, o_4, o_5, o_6\}.
\]
Note that the set includes
all outcomes whose inclusion is decided by stakeholders $1$ and $2$
on their own. Additionally, outcome $o_6$ is included because
$o_4\prefl_1 o_7$, $o_5\prefl_1 o_4$ and $o_6\prefl_2 o_5$. 
\label{ex:w1w2}
\end{example}
\begin{example}
For the induced preference graph in Figure~\ref{fig:induced-pref-graph}, 
consider evaluating the agreement condition. The set of outcomes that
are dominated by outcomes satisfying \texttt{No-Code} as per the stakeholder
$1$ is $\emptyset$. On the other hand, for stakeholder $2$, the set is
$\{o_5, o_7\}$ because $o_6\prefl_2 o_5$ and $o_3\prefl_2 o_7$. 

Therefore, for $\psi_2 = \texttt{No-Code}$,  
\[
\mathtt{A}_1: O\setminus \displaystyle\bigcup_{a\in \{1,2\}}\{o~|~\exists o'.o' \in \lsem{\psi_2}\ \land\ o \prefr_a^+ o'\} = O\setminus\{o_5, o_7\}.
\]
On the other hand, 
\[
\mathtt{A}_2: O\setminus \{o~|~\exists o'.o' \in \lsem{\psi_2}\ \land\ o \prefr_{\{1,2\}}^+ o'\}  = 
O\setminus \{o_4, o_5, o_7\}.
\]
The membership of
$o_4$ is decided from the relations: $o_6\prefl_2 o_5$  and $o_5\prefl_1 o_4$.
\label{ex:a1a2}
\end{example}

The combinations of $\mathtt{W}_1$ and $\mathtt{W}_2$ with $\mathtt{A}_1$ and $\mathtt{A}_2$  yield  four different
semantics for $\mlangpref{\psi_1}{\psi_2}{A}$. We will denote them by
$\lsem{\mlangpref{\psi_1}{\psi_2}{A}}{}^{\mathtt{W}_i\mathtt{A}_j}$ where
$i, j \in \{1,2\}$.

\begin{example}
Using the Examples~\ref{ex:w1w2} and \ref{ex:a1a2}, we have the following
when $\psi_1=\true$ and $\psi_2=\texttt{No-Code}$: 
\[
\begin{array}{llll}
\lsem{\mlangpref{\psi_1}{\psi_2}{\{1,2\}}}{}^{\mathtt{W}_1\mathtt{A}_2} & = &  \{o_1, o_2, o_3\}, \\

\lsem{\mlangpref{\psi_1}{\psi_2}{\{1,2\}}}{}^{\mathtt{W}_1\mathtt{A}_1} & = &  \{o_1, o_2, o_3, o_4\}, \\

\lsem{\mlangpref{\psi_1}{\psi_2}{\{1,2\}}}{}^{\mathtt{W}_2\mathtt{A}_2} & = & \{o_1, o_2, o_3, o_6\}, \\

\lsem{\mlangpref{\psi_1}{\psi_2}{\{1,2\}}}{}^{\mathtt{W}_2\mathtt{A}_1} & = &  \{o_1, o_2, o_3, o_4, o_6\}. 
\end{array}
\]
\label{ex:collab}
\end{example}

\noindent\textbf{Relationships Between Alternative Collaborative Semantics.} The following Theorem shows the relationship between the two witness conditions and the relationship between the two agreement conditions. 
\begin{theorem}\label{lemma:witness}
$\mathtt{W}_1 \subseteq \mathtt{W}_2$ and $\mathtt{A}_2 \subseteq \mathtt{A}_1$. 
\end{theorem}

\noindent\emph{Proof.\ }
(i) $\mathtt{W}_1 \subseteq \mathtt{W}_2$. \\
Consider any 
$o_1\in \mathtt{W}_1$. Then, $o_1 \in \bigcup_{a\in A}\{o~|~\exists o'.o' \in \lsem{\psi_2}\ \land\ o \prefl_a^+ o'\}$ by the definition of $\mathtt{W}_1$, 
Thus, there is an agent $a_1\in A$ and an outcome $o_2\in \lsem{\psi_2}{}$ such that $o_1\prefl_a^+ o_2$. Since $a\in A$, it then follows from the Definition~\ref{def:path} that 
$o_1 \prefl_A^+ o_2$. 
Therefore, $o_1\in \mathtt{W}_2$, by the definition of $\mathtt{W}_2$. 

(ii) $\mathtt{A}_2 \subseteq \mathtt{A}_1$.
We first show that 
\begin{equation}\label{2023-07-29}
\begin{split}
& \bigcup_{a\in A} \{o~|~\exists o'.o' \in \lsem{\psi_2}\ \wedge\ o \prefr_a^+ o'\}\\
\subseteq \, & \, 
\{o~|~\exists o'.o' \in \lsem{\psi_2}\ \wedge\ o \prefr_A^+ o'\}.
\end{split}
\end{equation}
For any $o_1\in \bigcup_{a\in A} \{o~|~\exists o'.o' \in \lsem{\psi_2}\ \wedge\ o \prefr_a^+ o'\}$,  there is an agent $a_1\in A$ and an outcome $o_2\in \lsem{\psi_2}{}$ such that $o_1\prefr_a^+ o_2$. Then, $o_1\prefr_A^+ o_2$ by Definition~\ref{def:path} because $a_1\in A$. Hence, statement~\eqref{2023-07-29} is true. 
Thus, it follow from the definitions of $\mathtt{A}_1$ and $\mathtt{A}_2$ that
\begin{equation*}
\begin{split}
\mathtt{A}_2 =\, & \{o~|~\forall o'.o' \in \lsem{\psi_2}\ \Rightarrow\ o \not\prefr_A^+ o'\}\\
= \, & \, O \setminus 
\{o~|~\exists o'.o' \in \lsem{\psi_2}\ \wedge\ o \prefr_A^+ o'\}\\
\subseteq \, & \, O \setminus 
\bigcup_{a\in A} \{o~|~\exists o'.o' \in \lsem{\psi_2}\ \wedge\ o \prefr_a^+ o'\}\\
= \, & \bigcap_{a\in A} \{o~|~\forall o'.o' \in \lsem{\psi_2}\ \Rightarrow\ o \not\prefr_a^+ o'\} = \mathtt{A}_1.  \phantom{AAAAA}\Box
\end{split}
\end{equation*}

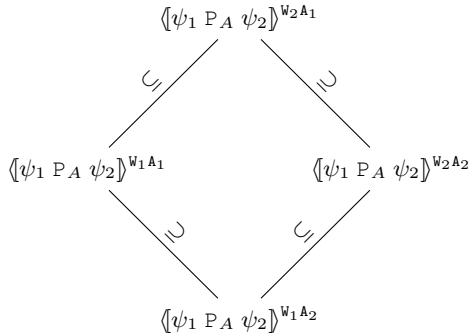
\begin{figure}[h]
\begin{center}
\begin{tikzpicture}[every edge quotes/.style = {auto, font=\footnotesize, sloped}]
\node[shape=rectangle,draw=white] (4) at (2,0) {$\lsem{\mlangpref{\psi_1}{\psi_2}{A}}{}^{\mathtt{W}_1\mathtt{A}_2}$};
\node[shape=rectangle,draw=white] (2) at (0,2) {$\lsem{\mlangpref{\psi_1}{\psi_2}{A}}{}^{\mathtt{W}_1\mathtt{A}_1}$};
\node[shape=rectangle,draw=white] (3) at (4,2) {$\lsem{\mlangpref{\psi_1}{\psi_2}{A}}{}^{\mathtt{W}_2\mathtt{A}_2}$};
\node[shape=rectangle,draw=white] (1) at (2,4) {$\lsem{\mlangpref{\psi_1}{\psi_2}{A}}{}^{\mathtt{W}_2\mathtt{A}_1}$};

\draw (2) edge["$\subseteq$"] (1)
      (3) edge["$\supseteq$"] (1)
      (4) edge["$\supseteq$"] (2)
      (4) edge["$\subseteq$"] (3);

\end{tikzpicture}
\end{center}
\caption{Relative Ordering of Semantics of Preference Queries }
\label{fig:relative}
\end{figure}

 The above theorem leads to the  relationship between different semantics of the query
 as illustrated in the Figure~\ref{fig:relative}.

\section{Answering Preference Queries}
\label{sec:mu-multi}
We now proceed to show how to answer multi-stakeholder preference queries.  Specifically, we show that  multi-stakeholder preference queries can be reduced to evaluating a corresponding alternation-free modal $\mu$-calculus expression. This allows us to take advantage of the state-of-the-art tools for $\mu$-calculus model-checking to efficiently answer multi-stakeholder preference queries.\\

\noindent\textbf{Modal $\mu$-calculus.} Modal $\mu$-calculus \cite{Koz83,EJS01}, ${\mathcal L}_{\mu}$, extends propositional modal logic by adding the least and the greatest fixed point operators.  $L_\mu$  uses explicit fixed point and modal operators to express temporal properties over events and states in a labeled transition system.  {\em Labeled transition systems} consist of a set of states,  a transition relation over state-pairs parameterized with events (transition annotations) and a labeling function that maps each state to a set of propositions that hold in that state. 
It is easy to see that an induced preference graph can be viewed as a labeled transition system over $O\cup\{\bot\}$, an annotated transition relation (edges being annotated with the set of stakeholders), and a labeling function mapping each outcome to the set of propositions satisfied by the outcome. The primary difference is that the edge-annotation is a set (in an induced preference graph) rather than a symbol (in a labeled transition systems). Note, however, that such a difference is purely syntactical as we can replace an edge annotated with a set by a set of edges, where each edge in the set is annotated by a distinct member of the set. 
We will use `states' and `outcomes' interchangeably in refering to an induced preference graph interpreted as a labeled transition system. \\

\noindent\textbf{Syntax of Modal $\mu$-calculus. } The syntax of $\mu$-calculus involves propositional constants, atomic propositions, modalities, fixed
point variables and expressions and Boolean connectives: 
\[
\small
\begin{array}{lr}
\phi \rightarrow & \true~|\false~|~\texttt{AP}~|~\neg\phi~|~\phi\land\phi~|~\phi\lor\phi~|~\moddiam{A}\phi
~|~Z~|~\mu Z.\phi
\end{array}
\]

In the above, the parameter $A$ of the modal operator ($\moddiam{A}$) is associated with the edge annotation of the labeled transition system.  In an induced preference graph, each edge is annotated with a subset of all stakeholders. In our context, in the modal operators, $A$ will represent a set of stakeholders. When $A$ is singleton such as $A=\{a\}$, we will denote the modal condition as $\moddiam{a}$.\\

\noindent\textbf{Semantics of Modal $\mu$-calculus. } The semantics of $\mu$-calculus formula is given in terms of a set of
states in a labeled transition system that satisfy the formula.
The semantics is specified by the function  $\msem{\ }{}:\Phi\times
\mathcal{E}\times \mathcal{\ipg}\rightarrow \mathcal{P}(O)$ where
$\mathcal{E}$ is the power set of mappings of fixed point variables to
outcomes in $O$. This mapping is referred to as the environment:
$e:\mathcal{Z} \rightarrow \mathcal{P}(O)$; $\mathcal{Z}$ being the
set of fixed point variables in the formula whose semantics is being
evaluated. We will use the notation $e[Z\mapsto O']$ to denote the
environment where the mapping of fixed point variable $Z$ in $e$ is
updated to $O'\subseteq O$. We omit $\ipg$ when it is not necessary to distinguish between different induced preference graphs.

Figure \ref{fig:musem} shows the semantics of $\mu$-calculus. The propositional constants $\true$ and $\false$ are satisfied by all states and no states, respectively. The atomic proposition $p$ is satisfied in all states whose labeling includes $p$. The formula
$\varphi_1\land\varphi_2$ is satisfied by all states that satisfy both $\varphi_1$ and $\varphi_2$. The formula $\moddiam{A}\varphi$ is satisfied by any state which has at least one next state (reachable via an edge annotated with a set that has a non-empty intersection with $A$) that satisfies $\varphi$. 

\begin{figure}[t]
\small
\[
\begin{array}{r@{\extracolsep{-0.01em}}cl}
\msem{\true}{e} & = & O \cup \{\bot\} \ \ \ \ \   \msem{\false}{e} \ = \ \emptyset  \ \ \ \ \  
\msem{p}{e} \ = \ \{o~|~p\in L(o)\} \\[0.5em]

\msem{\neg\varphi}{e} & = & O - \msem{\varphi}{e} \ \ \ \ \ 
\msem{\varphi_1\land\varphi_2}{e} \ = \ \msem{\varphi_1}{e} \ \cap\ \msem{\varphi_2}{e} \\[0.5em]

\msem{\moddiam{A}\varphi}{e} & = & \{s~|~\exists s'. ((A\cap A'\neq \emptyset) \ \land\ s\trans{A'} s')\ \land\ s' \in \msem{\varphi}{e}\} \\[0.5em]


\msem{Z}{e} & = & e(Z) \\ [0.5em]

\msem{\mu Z.\varphi}{e} & = & f^{|O|}_{Z,\varphi,e}(\emptyset) 

\mbox{ where } f_{Z,\varphi,e}(O') = \msem{\varphi}{e[Z\mapsto O']} \mbox{ and } O'\subseteq O

\end{array}
\]
\caption{$\mu$-calculus Semantics}
\label{fig:musem}
\end{figure}

The semantics of fixed
point variable $Z$ is given by the environment mapping $e$.  The
semantics of least fixed point formula $\mu Z.\varphi$ is computed by the $|O|$ applications of function $f_{Z, \varphi, e}$ on $\emptyset$
(Tarski-Knaster fixed point theorem~\cite{Tarski}). We omit the greatest fixed point construct as its semantics can be realized using the least fixed point and negation. 

Model checking a labeled transition system against
a given $\mu$-calculus formula amounts to identifying
the set of states in the transition system that
belong to the semantics of the $\mu$-calculus formula.
\\

\noindent\textbf{Alternation-Free Modal $\mu$-calculus. }For our purposes, it turns out that we only need the  alternation-free  fragment ${\mathcal L}^{af}_{\mu}$ \cite{emerson1986efficient} of ${\mathcal L}_\mu$. An attractive property of ${\mathcal L}^{af}_{\mu}$ is that in it there is no real interaction between least and greatest fixpoint operators \cite{marti2021focus}, which, at the expense of reduced expressive power relative to ${\mathcal L}_\mu$, yields more efficient reasoning  \cite{marti2021focus,EJS01}.\\

\noindent\textbf{Translating Query Language to $\mu$-calculus. }
We present a strategy to evaluate the proposed preference queries using model checking. We will augment the induced preference graph which encodes a labeled transition system with additional reverse edges; this will help in explaining the answers to multi-stakeholder preference queries in relation to the stakeholder preferences and the chosen semantics; however, in the implementation, such reverse edges can be handled implicitly.  For every edge from $o_j$ to $o_i$ due to preference $o_i\prefl_a o_j$ of stakeholder $a$,  we will add a reverse edge from $o_i$ to $o_j$. 

\begin{figure*}[t]
\small
\[
\begin{array}{rcl}
Tr^{t}(X) & = & X \hfill \mbox{if } X \mbox{is proposition or propositional constants} \\[1em]

Tr^{t}(\neg \psi) & = & \neg Tr^{t}(\psi) \\[1em]

Tr^{t}(\psi_1\ b\ \psi_2) & = & Tr^{t}(\psi_1) \ b\ Tr^{t}(\psi_2) \hfill\mbox{where } b \in \{\land,\lor\}\\[1em]

Tr^{t}(\mlangpref{\psi_1}{\psi_2}{A}) & = &
\left\{
\begin{array}{ll}
Tr^{cs}(\psi_1) \ \land \ \displaystyle\bigwedge_{a\in A} \left[(\mu Z.(\moddiam{a}_r Tr^{cs}(\psi_2) \ \lor \ \moddiam{a}_rZ)) \ \land\
(\neg \mu Z.(\moddiam{a} Tr^{cs}(\psi_2) \ \lor\ \moddiam{a}Z))\right] & \mbox{if } t=cs \\[0.5em]

Tr^{\mathtt{W}_1\mathtt{A}_2}(\psi_1)\  \land\  \displaystyle\bigvee_{a\in A} \left(\mu Z.(\moddiam{a}_r Tr^{\mathtt{W}_1\mathtt{A}_2}(\psi_2)\ \lor\ \moddiam{a}_rZ)\right) 
  \ \land\  \neg \mu Z.(\moddiam{A}Tr^{\mathtt{W}_1\mathtt{A}_2}(\psi_2)\ \lor\ \moddiam{A}Z) & \mbox{if } t=\mathtt{W}_1\mathtt{A}_2\\[0.5em]

Tr^{\mathtt{W}_1\mathtt{A}_1}(\psi_1)\  \land\  \displaystyle\bigvee_{a\in A} \left(\mu Z.(\moddiam{a}_r Tr^{\mathtt{W}_1\mathtt{A}_1}(\psi_2)\ \lor\ \moddiam{a}_rZ)\right) \ \land\ \displaystyle\bigwedge_{a\in A} \left(\neg \mu Z.(\moddiam{a} Tr^{\mathtt{W}_1\mathtt{A}_1}(\psi_2)\ \lor\ \moddiam{a}Z)\right)  
& \mbox{if } t = \mathtt{W}_1\mathtt{A}_1 \\[0.5em]

Tr^{\mathtt{W}_2\mathtt{A}_2}(\psi_1)\   \land\ \mu Z.(\moddiam{A}_r Tr^{\mathtt{W}_2\mathtt{A}_2}(\psi_2)\ \lor\ \moddiam{A}_rZ) \
 \land\  \neg\mu Z.(\moddiam{A}Tr^{\mathtt{W}_2\mathtt{A}_2}(\psi_2)\ \lor\ \moddiam{A}Z) & \mbox{if } t = {\mathtt{W}_2\mathtt{A}_2} \\[0.5em]

 Tr^{\texttt{W2A1}}(\psi_1)\  \land\ \mu Z.(\moddiam{A}_r Tr^{\mathtt{W}_2\mathtt{A}_1}(\psi_2)\ \lor\ \moddiam{A}_rZ) \
\land\ \displaystyle\bigwedge_{a\in A} \left(\neg \mu Z.(\moddiam{a} Tr^{\mathtt{W}_2\mathtt{A}_1}(\psi_2)\ \lor\ \moddiam{a}Z)\right)
& \mbox{if } t = {\mathtt{W}_2\mathtt{A}_1 }
\end{array}
\right.
\end{array}
\]
\caption{Translation of Multi-Stakeholder Preference Queries into $\mu$-calculus}
\label{fig:translation}
\end{figure*}

Therefore, the set $\{o~|~\exists o'.o' \in \lsem{\psi}\ \land\ o \prefl^+_a o'\}$
can be expressed in $\mu$-calculus as:
$
\mu Z.(\moddiam{a}_r\psi \ \lor \ \moddiam{a}_r Z)
$.
The semantics captures the set of states which can reach some state
satisfying $\psi$ via one or more reverse edges; the modal requirement
$\moddiam{a}_r$ is satisfied using reverse edges annotated with $a$.

\begin{example}
Consider the formula 
$\mu Z. (\moddiam{1}_r\texttt{Code} \ \lor \ \moddiam{1}_r Z)$ representing the set of all states that have a path to a state satisfying \texttt{Code} via one or more edges annotated with $1$. 
We evaluate this expression using the induced preference graph shown in Figure~\ref{fig:induced-pref-graph}. 

Let $\varphi$ denotes $(\moddiam{1}_r\texttt{Code} \ \lor \ \moddiam{1}_r Z)$. 
Therefore,  $\msem{\mu Z.\varphi}{e} = f^{8}_{Z,\varphi,e}(\emptyset)$ where $f_{Z,\varphi,e}(O') = \msem{\varphi}{e[Z\mapsto O']}$.
\[
\begin{array}{l}
f_{Z,\varphi,e}(\emptyset) = \msem{\moddiam{1}_r\texttt{Code} \ \lor \ \moddiam{1}_r Z}{e[Z\mapsto \emptyset]} \\ [1em]

= \msem{\moddiam{1}_r\texttt{Code}}{e[Z\mapsto\emptyset]} \ \cup\ \msem{\moddiam{1}_r Z}{e[Z\mapsto \emptyset]} \\ [1em]

= \{o~|~\exists o'.(o\trans{A}_r o'\ \land\ A \cap \{1\} \neq \emptyset)\ \land\ o' \in \msem{\texttt{Code}}{e[Z\mapsto \emptyset]}\} \\
\ \ \cup \\ 
\ \ \ \ \{o~|~\exists o'.(o\trans{A}_r o'\ \land\ A \cap \{1\} \neq \emptyset)\ \land\ o' \in \msem{Z}{e[Z\mapsto \emptyset]}\} \\
\mbox{where} \trans{A}_r \mbox{ denotes reverse edge relations} \\ [1em]

= \{(o\trans{A}_r o'\ \land\ A \cap \{1\} \neq \emptyset)\ \land\ o'\in \{o_1,o_2,o_4,o_5\}\} \\
\ \ \cup \\
\ \ \ \ \{o~|~\exists o'.(o\trans{A}_r o'\ \land\ A \cap \{1\} \neq \emptyset)\ \land\ o' \in \emptyset\} \\[1em] 

= \{o_1,o_5\} \ \cup\ \emptyset = \{o_1, o_5\}
\end{array}
\]
Proceeding further
\[
\begin{array}{l}
f^2_{Z,\varphi,e}(\emptyset) = f_{Z, \varphi, e}(f_{Z, \varphi, e}(\emptyset) = f_{Z,\varphi,e}(o_1, o_5) \\[1em]

= \msem{\moddiam{1}_r\texttt{Code}}{e[Z\mapsto\{o_1,o_5\}]} \ \cup\ \msem{\moddiam{1}_r Z}{e[Z\mapsto \{o_1,o_5\}]} \\[1em]

= \{o~|~\exists o'.(o\trans{A}_r o'\ \land\ A \cap \{1\} \neq \emptyset)\ \land\ o' \in \msem{\texttt{Code}}{e[Z\mapsto \{o_1,o5\}]}\} \\
\ \ \cup \\ 
\ \ \ \ \{o~|~\exists o'.(o\trans{A}_r o'\ \land\ A \cap \{1\} \neq \emptyset)\ \land\ o' \in \msem{Z}{e[Z\mapsto \{o_1,o_5\}]}\} \\ 
\mbox{where} \trans{A}_r \mbox{ denotes reverse edge relations} \\ [1em]
\end{array}
\]

\[
\begin{array}{l}
= \{(o\trans{A}_r o'\ \land\ A \cap \{1\} \neq \emptyset)\ \land\ o'\in \{o_1,o_2,o_4,o_5\}\} \\
\ \ \cup \\
\ \ \ \ \{o~|~\exists o'.(o\trans{A}_r o'\ \land\ A \cap \{1\} \neq \emptyset)\ \land\ o' \in \{o_1,o_5\}\} \\[1em]

= \{o_1,o_5\} \ \cup \emptyset = \{o_1,o_5\} 
\end{array}
\]
The (least) fixed point is reached as further application of $f$ onto itself will not alter the
result.

\label{ex:mu-single}
\end{example}

Similarly, the set $\{o~|~\forall o'.o' \in \lsem{\psi}\ \Rightarrow\ o \not\prefr^+_a o'\}$
is equal to $O\setminus \{o~|~\exists o'.o' \in \lsem{\psi}\ \land\ o \prefr^+_a o'\}$,
which  can be expressed in $\mu$-calculus as:
$
\neg\mu Z.(\moddiam{a}\psi\ \lor\ \moddiam{a}Z)
$.
This semantics yields the set of states which have no path to any state that satisfies $\psi$.
Hence, a query of the form $\mlangpref{p}{q}{a}$, where $p$ and $q$ are atomic
propositions, can
be expressed in $\mu$-calculus as:
\[
p \ \land \ \left[\mu Z.(\moddiam{a}_r q \ \lor \ \moddiam{a}_rZ)\right] \ \land\
\left[\neg\mu Z.(\moddiam{a}q\ \lor\ \moddiam{a}Z) \right]
\]

Now, in \textit{\collaba\ Collaboration},
the set of outcomes that dominate at least one outcome satisfying $\psi$
for a set $A$ of stakeholders is given by
$
\bigcup_{a\in A} \{o~|~\exists o'.\ o' \in \psi\ \land \ o\prefl_a^+ o'\}
$
which in turn is reflected by the semantics of the $\mu$-calculus formula:
$
\bigvee_{a\in A} \left(\mu Z.(\moddiam{a}_r\psi\ \lor\ \moddiam{a}_rZ)\right)
$.
The preceding formula identifies the set of outcomes that have path(s) to some outcome satisfying $\psi$ in the transpose-induced preference graph (i.e., using reversed edges)
$\ipg$. Along each path that decides reachability, each of the edges must be 
annotated by the same $a$.

On the other hand, in the \textit{\collabb\ Collaboration}, the domination of outcomes over at least one outcome satisfying
$\psi$ for a set $A$ of stakeholders is decided by
$
\{o~|~\exists o'.\ o' \in \psi\ \land \ o\prefl_A^+ o'\}
$
which in turn corresponds to the semantics of the $\mu$-calculus formula
$
\mu Z.(\moddiam{A}_r\psi\ \lor\ \moddiam{A}_rZ)
$.
This denotes the set of outcomes that have path(s) to some outcome satisfying $\psi$ in the transpose-induced preference graph $\ipg$; the reachability
is determined by the edges annotated by at least one
element from $A$.

Thus, the Witness and Agreement Conditions can be
expressed in $\mu$-calculus as follows:
\[
\begin{array}{rl}
\mathtt{W}_1: & \mbox{semantics of } \displaystyle\bigvee_{a\in A} \left(\mu Z.(\moddiam{a}_r\psi\ \lor\ \moddiam{a}_rZ)\right) \\[1em]
\mathtt{W}_2: & \mbox{semantics of }  \mu Z.(\moddiam{A}_r\psi\ \lor\ \moddiam{A}_rZ) \\[1em]
\mathtt{A}_1: & \mbox{semantics of }  \displaystyle\bigwedge_{a\in A} \left(\neg\mu Z.(\moddiam{a}\psi\ \lor\ \moddiam{a}Z)\right) \\[1em]
\mathtt{A}_2: & \mbox{semantics of }  \neg\mu Z.(\moddiam{A}\psi\ \lor\ \moddiam{A}Z) \\
\end{array}
\label{eq:WitAgreemucalc}
\]

Figure~\ref{fig:translation} shows the translation function that, given an expression in the multi-stakeholder preference query language and the chosen multi-stakeholder preference semantics as arguments, outputs the corresponding ${\mathcal L}^{af}_{\mu}$  expression. The run-time for translation is linear in the size ($n$) of the number of operators ($\land,\lor,\neg,\mathtt{P}$) in the query. The size of the translation is of the order $O(|\mathcal{A}|^k\times n^k)$, where
$|\mathcal{A}|$ is the number of stakeholders, $k$ the \emph{nesting depth} of the queries of the form $\mlangpref{\psi_1}{\psi_2}{A}$ and $n$ the size of the query. For instance, for a query of the form $(\mlangpref{p}{(\mlangpref{q}{r}{B})}{A})$, $n$ and $k$ are both equal to $2$.  The run-time for model checking ${\mathcal L}^{af}_{\mu}$ formula is linear in the size of the formula and the state space of the labeled transition
system (induced preference graph). We expect  the nesting depth of the query to be reasonably small and the run-time will be determined largely by the number of stakeholders in the query and the
size of the number of outcomes (size of the induced preference graph). Note, however, that the number of outcomes is exponential in the number of attributes describing the outcomes, as in the case of reasoning with qualitative preferences \cite{Goldsmith:JAIR08}. 

The following theorem establishes the correctness of reduction of multi-stakeholder preference queries to ${\mathcal L}^{af}_{\mu}$ expressions.
\begin{theorem}
For a multi-stakeholder preference query $\psi$ 
(as described in Section \ref{sec:multis}), $o\in\lsem{\psi}{\ipg}^{i}$
if and only if $o\in\msem{Tr^i(\psi)}{\ipg}$, where $\ipg$
is the preference graph induced by the stakeholder preferences and $i$ denotes
the type (consensus or variants of collaborative) of semantics used to answer
$\psi$. 
\label{thm:singlecorrectness}
\end{theorem}

The proof of Theorem \ref{thm:singlecorrectness} proceeds by induction over the structure of the mult-stakeholder preference query. 

\section{Implementation}
\label{sec:impl}

We have implemented a multi-stakeholder preference reasoner in XSB tabled logic programming environment~\cite{swift2012xsb} to demonstrate the viability of
our approach. The logical encoding of ${\mathcal L}^{af}_{\mu}$ used allows for \emph{on-the-fly} evaluation of logical queries, circumventing the need for constructing the complete multi-stakeholder induced preference graph. In other words, only the portion of the induced preference graph relevant for answering the query is constructed, resulting in significant savings in computational and memory savings relative to a naive implementation. 

\subsection{Input: Preferences as Logical Relations and Facts}
\label{sec:input}

The implementation takes as input a XSB Prolog file containing (a) preference
specifications described in terms of "flips" relation and (b) 
logical fact specifying the different valuations of the attributes
that describe each outcome. 

In the following, we describe the representation of 
preferences of each agents as logical relations in XSB Prolog.
Consider that there are $n$ attributes $x_1, x_2, \ldots, x_n$
that describe the outcomes, where valuations of $x_i$ 
are $v_{i_1}, v_{i_2}, \ldots, v_{i_k}$. This is 
captured by a Prolog fact:

{\footnotesize
\begin{verbatim}
properties([v11, v12, v13, ..., vik],
            [v21, v22, v23, ..., v2k],
            ...
            [vn1, vn2, vn3, ..., vnk]).
\end{verbatim}
}
Note that, the each argument of the term is a Prolog list
is the domain of the corresponding attribute; $i^{th}$ argument
being the domain of the valuations of $x_i$.

Next, let agent $a$ has the preference
\[
[x_1=v_1]\ x_2=v_2 \prec x_2=v'_2\ [x_3, x_4]
\]
capturing the fact that when $x_1=v_1$, any outcome $o'$
where $x_2=v'_2$ is preferred to outcomes $o$ where
$x_j=v_2$ regardless of the valuations of
$x_3$ and $x_4$ (all other attribute valuations in
$o$ and $o'$ being same). 

This is represented by the logical
relation

{\footnotesize
\begin{verbatim}
trans([v_1, v_2, _, _, _V5, _V6, ..., _Vn],
       a,
       [v_1, v_2', _, _, _V5, _V6, ..., _Vn]).
\end{verbatim}
}

In the above, the first, second and third arguments of the \texttt{trans} relation captures the outcomes $o$, agent
$a$ and outcomes $o'$, respectively, Note that, the
valuation of $x_1$ in both $o$ and $o'$ are $v_1$;
the valuations of $x_3$ and $x_4$ are captured by
"don't care" logical variables ("\texttt{\_}") indicating they
can be any valuation;  the valuation of $x_5$ till
$x_n$ are are any values that are same in both $o$
and $o'$.

\begin{example}
In Figure~\ref{fig:pref-statements}, the preference for agent $1$
\[
P_1\ \ E = Code \succ_E E = No\_Code
\]
is captured by the logical relation 

{\footnotesize
\begin{verbatim}
trans([noCode, _X, _Y], 1, [code, _X, _Y]).
\end{verbatim}
}

The relation captures the fact that all else being equal, any
outcome with first attribute valuation \texttt{code} is preferred
to any outcome with first attribute valuation \texttt{noCode}. 

This input file also contains the logical fact:

{\footnotesize
\begin{verbatim}
properties([code, noCode], 
            [simple, complex], 
            [fix, noFix]).
\end{verbatim}
}

which states that the outcomes being considered
contain three attributes and the valuation of three
attributes are given in the form of XSB Prolog list.
\end{example}

The listing for the preferences in Figure~{\ref{fig:pref-statements}}
in (XSB) Prolog is as follows:

{\footnotesize
\begin{verbatim}
trans([noCode, _X, _Y], S, [code, _X, _Y]) 
     :- S = 1; S = 3.

trans([code, _X, fix], 1, [code, _X, noFix]).

trans([noCode, _X, fix], 2, [code, _X, fix]).

trans([_X, complex, fix], 2, [_X, simple, fix]).

trans([code, complex, noFix], 2, 
    [noCode, simple, noFix]).

trans([_X, complex, _Y], 3, [_X, simple, _Y]).

trans([noCode, _, _], 3, [code, _, _]).
\end{verbatim}
}

\subsection{Modules of Prototype Implementation}
\label{sec:modules}

There are three primary modules in the implementation:
a module for $\mu$-calculus model checker, a module for
for appropriately translating the preference queries
to $\mu$-calculus formula, which is used in the
module to evaluate the semantics of preference query. 

The model checker is written using tabling in
XSB Prolog (that allows for efficient least fixed point
computation) in less than 100 lines of Prolog code. 
The relation
{\footnotesize
\begin{verbatim}
models(S, Phi)
\end{verbatim}
}
returns true when the instantiation of variable \texttt{S}
to some state of a Kripke structure satisfies
the $\mu$-calculus formula captured by the variable \texttt{Phi}.
This is a local, on-the-fly realization of the semantics
of $\mu$-calculus formula, where the state-space of
Kripke structure is explored only if it is necessary to
prove the satisfiability of \texttt{Phi} at state \texttt{S}.

To illustrate the connection between the \texttt{models}
relation and the \texttt{trans} relation describing
the preferences, we present below the definition of
\texttt{models} for $\moddiam{.}$-modal formulas. 
{\footnotesize
\begin{verbatim}
models(S, diam(A, Phi)) :-
    trans(S, A1, S1),
    member(A1, A), 
    models(S1, Phi).
\end{verbatim}
}
The above definition states that if there exists
a \texttt{trans} relation over \texttt{S}, \texttt{A1}
and \texttt{S1} indicating some stackholder \texttt{A1} prefers
outcome \texttt{S1} over outcome \texttt{S}; if
\texttt{A1} is a member of \texttt{A}; and if 
\texttt{S1} satisfies the formula \texttt{Phi}, then
we conclude that \texttt{S} satisfies \texttt{diam(A, Phi)}
($\moddiam{A}\varphi$).

The module \texttt{translate} ($<100$ lines of XSB Prolog) contains the definition of
{\footnotesize
\begin{verbatim}
translate(F1, Type, F2)
\end{verbatim}
}
where \texttt{F1} is the formula in preference query
language, \texttt{F2} is the corresponding formula
in the $\mu$-calculus and \texttt{Type} captures
the different combinations of witness and agreement
conditions used in the translation. 

Finally, the module for computing the semantics of 
the query language includes the definition of
{\footnotesize
\begin{verbatim}
sem(F, Type, R)
\end{verbatim}
}
where \texttt{F} is the query, \texttt{Type}
is the combination of witness and agreement conditions
to be used to evaluate the query and \texttt{R} is
the result of the query. For instance,
{\footnotesize
\begin{verbatim}
sem(p(Psi1, Psi2, A), Type, R) :-
    sem(Psi1, Type, L),
    translate(p(true, Psi2, A), Type, MuForm),
    models_list(L, MuForm, R).
\end{verbatim}
}
presents the following computation for the 
formula \texttt{p(Psi1, Psi2, A)} (representing
the query $\mlangpref{\psi_1}{\psi_2}{A}$) . First, we
compute the semantics of \texttt{Psi1}, the result
of which is captured in \texttt{L}. That is,
\texttt{L} is the list of outcomes that
satisfy \texttt{Psi1}. Next, we translate
\texttt{p(true, Psi2, A)} to the corresponding
$\mu$-calculus formula \texttt{MuForm}. Finally, we identify
the outcomes in \texttt{L} that satisfiy
\texttt{MuForm} and include them in \texttt{R}. 
This is performed by the \texttt{models\_list} predicate
which calls the \texttt{models} predicate (see
$\mu$-calculus model checker module above) on
each element of \texttt{L}.

\begin{figure*}[h]
{\scriptsize
\begin{verbatim}
| ?- sem(p(true, prop(noCode), [1,2]), w1a2, R).

R = [[code,simple,fix],[code,simple,noFix],[noCode,simple,fix]];

no
| ?- sem(p(true, prop(noCode), [1,2]), w1a1, R).

R = [[code,simple,fix],[code,simple,noFix],[code,complex,fix],[noCode,simple,fix]];

no
| ?- sem(p(true, prop(noCode), [1,2]), w2a2, R).

R = [[code,simple,fix],[code,simple,noFix],[noCode,simple,fix],[noCode,simple,noFix]];

no
| ?- sem(p(true, prop(noCode), [1,2]), w2a1, R).

R = [[code,simple,fix],[code,simple,noFix],[code,complex,fix],[noCode,simple,fix],[noCode,simple,noFix]];

no


\end{verbatim}
}
\caption{Evaluation of the semantics of query $(\mlangpref{\true}{no\_Code}{A})$ with different types of collaboration}
\label{fig:mu-collab}
\end{figure*}

\subsection{On-the-fly Evaluation of Logical Statements}

It is worth noting that logical encoding allows
for on-the-fly evaluation. Intuitively, this
means the a query of the form:
{\footnotesize
\begin{verbatim}
    sem(p(true, prop(noCode), [1,2]), w1a2, R)
\end{verbatim}
}
is resolved only by considering the
trans-predicates that are related to 
stakeholders 1 and 2, and by considering
only those trans-predicates that are necessary
for the resolution. For instance, when the
above query eventually requires the resolution
of {\footnotesize \texttt{
models([code,simple,fix],rdiam([1],prop(noCode)))}}
our implementation will try to resolve the
predicate {\footnotesize\texttt{trans(X,1,[code,simple,fix])}},
by finding the valuation of  \texttt{X} for which the above predicate is true. 
Note that, we are not exploring all the transition relations for all stakeholders and for all outcomes. There may be multiple valuations for \texttt{X}; the logical encoding will find one of them and try to answer {\footnotesize\texttt{models([code,simple,fix],rdiam([1],prop(noCode)))}}. If the answer is false, then the encoding will consider another valuation for \texttt{X}; otherwise, it will not explore any other solutions for \texttt{X}. 

In short, the entire induced preference graph for all stakeholders is never constructed and the exploration proceeds by considering only the edges that are necessary for answering a query.  

\subsection{Evaluating Queries}

The listing of queries from Example~\ref{ex:collab} is presented in Figure~\ref{fig:mu-collab}.

\section{Preliminary Experiments}

To stress-test our implementation, we conducted two
types of experiments. For the first type of experiments, we generated for each stakeholder, random preference statements over $n$ binary preference variables, while ensuring to disallow inconsistent preferences. The resulting preferences statements include direct preferences between outcomes
described by the attribute values, conditional preferences
between attribute values, and relative importance between attributes.  The results of this set of experiments are summarized in Table~\ref{tbl:randompref}.
Table entries show the run-time (in seconds) for answering some representative multi-stakeholder preference queries based on different choices of semantics, for several choices of the number of attributes.  The numbers
in parenthesis indicate the size of the solution set for the corresponding query. 

The results in Table~\ref{tbl:randompref} shows the viability of
our approach; in each of the 84 cases, the corresponding query is answered in 
at most 2 seconds. Recall that the run-time for answering a query depends on the
nesting depth of the query, the number of stakeholders that appear in the query, and
the size of the  preference graph induced by their preferences. We observe that
the run-time for answering queries for semantic type $\texttt{W}_1\texttt{A}_1$ is the smallest, whereas that for semantic type
$\texttt{W}_2\texttt{A}_2$ is the
largest. This is explained by the fact that both the
witness and agreement conditions in the case of the former are evaluated using disjunctive constraints, whereas in the case of the latter, they are evaluated using chaining constraints. This implies the state space explored for the latter is at least large as the state space explored for the former.

\begin{table}[t]
\scriptsize
\begin{tabular}{c|c|lll}
\multirow{2}{*}{\bf Query} & \multirow{2}{*}{\bf Type} & \multicolumn{3}{c}{\bf Number of Attributes} \\ \cline{3-5} 
& & 5 & 6 &  8 \\ \hline\hline
\multirow{2}{*}{$\mlangpref{tt}{(\mlangpref{tt}{tt}{L1})}{L2}$} 
       & $\texttt{W}_1\texttt{A}_2$ & 0.02 (2) & 0.23 (10) & 0.27 (0)\\
       & $\texttt{W}_1\texttt{A}_1$ & 0.02 (2) & 0.11 (10) & 0.12 (0) \\
\multirow{2}{*}{$L1=\{1,2\}, L2=\{3,4\}$} 
       & $\texttt{W}_2\texttt{A}_2$ & 0.02 (2) & 0.41 (12) & 0.49 (0) \\
       & $\texttt{W}_2\texttt{A}_1$ & 0.02 (2) & 0.31 (12)  & 0.29 (0) \\ \hline

\multirow{2}{*}{$\mlangpref{tt}{(\mlangpref{tt}{tt}{L1})}{L2}$} 
       & $\texttt{W}_1\texttt{A}_2$ & 0.02 (2) & 0.28 (0) & 0.25 (32)\\
       & $\texttt{W}_1\texttt{A}_1$ & 0.02 (5) & 0.14 (0) & 0.16 (32)\\
\multirow{2}{*}{$L1=\{2,3\}, L2=\{4, 5\}$} 
       & $\texttt{W}_2\texttt{A}_2$ & 0.02 (4) & 0.39 (0) &  0.35 (32)\\
       & $\texttt{W}_2\texttt{A}_1$ & 0.02 (5) & 0.27 (0) & 0.27 (32) \\ \hline

\multirow{2}{*}{$\mlangpref{tt}{(\mlangpref{tt}{tt}{L1})}{L2}$} 
       & $\texttt{W}_1\texttt{A}_2$ & 0.01 (0) & 0.13 (6) & 0.02 (0) \\
       & $\texttt{W}_1\texttt{A}_1$ & 0.01 (0) & 0.07 (12) & 0.02 (0) \\
\multirow{2}{*}{$L1=\{5,6\}, L2=\{9, 10\}$} 
       & $\texttt{W}_2\texttt{A}_2$ & 0.01 (0) & 0.24 (6) & 0.02 (0) \\
       & $\texttt{W}_2\texttt{A}_1$ & 0.01 (0) & 0.17 (12) & 0.02 (0) \\ \hline

\multirow{2}{*}{$\mlangpref{tt}{(\mlangpref{tt}{tt}{L1})}{L2}$} 
       & $\texttt{W}_1\texttt{A}_2$ & 0.08 (1) & 0.54 (0) & 0.91 (23) \\
       & $\texttt{W}_1\texttt{A}_1$ & 0.05 (1) & 0.23 (0)  & 0.34 (27)\\
\multirow{2}{*}{$L1=\{1,2,3\}, L2=\{4,5,6\}$} 
       & $\texttt{W}_2\texttt{A}_2$ & 0.29 (2) & 1.03 (0) &  1.76 (36) \\
       & $\texttt{W}_2\texttt{A}_1$ & 0.21 (3) & 0.63 (0) & 0.95 (42) \\ \hline

\multirow{2}{*}{$\mlangpref{tt}{(\mlangpref{tt}{tt}{L1})}{L2}$} 
       & $\texttt{W}_1\texttt{A}_2$ & 0.04 (2) & 0.48 (0) &  0.46 (28)\\
       & $\texttt{W}_1\texttt{A}_1$ & 0.03 (2) & 0.23 (0) & 0.23 (29)\\
\multirow{2}{*}{$L1=\{2,3, 4\}, L2=\{5,6,7\}$} 
       & $\texttt{W}_2\texttt{A}_2$ & 0.09 (2) & 0.82 (0) &  0.82 (33)  \\
       & $\texttt{W}_2\texttt{A}_1$ & 0.07 (2) & 0.45 (0) &  0.52 (36)\\ \hline

\multirow{2}{*}{$\mlangpref{tt}{(\mlangpref{tt}{(\mlangpref{tt}{tt}{L1})}{L2})}{L3}$} 
       & $\texttt{W}_1\texttt{A}_2$ & 0.23 (0) & 1.16 (0) & 0.91 (0)\\
       & $\texttt{W}_1\texttt{A}_1$ & 0.19 (0) & 0.76 (0) & 0.65 (0)\\
$L1=\{1,2,3\}, L2=\{4,5,6\},$
       & $\texttt{W}_2\texttt{A}_2$ & 0.70 (2) & 1.73 (0) & 1.58 (0)  \\
$ L3=\{7,8,9\}$        & $\texttt{W}_2\texttt{A}_1$ & 0.54 (2)  & 1.27 (0)  & 1.02 (0)\\ \hline

\multirow{2}{*}{$\mlangpref{true}{(\mlangpref{tt}{(\mlangpref{tt}{tt}{L1})}{L2})}{L3}$} 
       & $\texttt{W}_1\texttt{A}_2$ & 0.26 (0) & 1.11 (0) & 0,76 (0)\\
       & $\texttt{W}_1\texttt{A}_1$ & 0.24 (0) & 0.72 (0) & 0.66 (0) \\
$L1=\{2,3,4\}, L2=\{5,6,7\},$
       & $\texttt{W}_2\texttt{A}_2$ & 0.65 (2)  &  1.54 (0) & 1.69 (0) \\
$ L3=\{8,9,10\}$        & $\texttt{W}_2\texttt{A}_1$ 
& 0.58 (2) &  1.14 (0) & 1.29 (0) \\ \hline

\end{tabular}
\caption{Experiments with Randomly Generated Preference Statements}
\label{tbl:randompref}
\end{table}

For the second set of experiments, we randomly generated
graphs in which the vertices correspond to  outcomes, and the edges denote preference between pairs of outcomes.
Note that in this case, because the graphs are randomly generated, and not induced by the stakeholder preferences, it is possible for the preferences reflected in the graph to be inconsistent, i.e., individual stakeholder's preference graph may be inconsistent. Each edge is annotated with a random subset of stakeholders (simulating
the setting where the stakeholder preferences induce edges in the induced preference graph).  
Table~\ref{tbl:timing} presents the timing results of our experiments
with random graphs. 

The column "configuration"
includes three numbers describing the randomly
generated induced preference graph: the first
number is the number of stakeholders, the
second number is the number of outcomes and
the third number indicates that the maximum
number of edges per stakeholder in
the induced preference graph.  We generate 
$25$ graphs per configuration. For each
configuration, we compute the result
of the three types of queries presented
in the first row of the table. For
each query, we consider four different
collaborative semantics and report the
time in seconds needed for the computation.

Typically, as the induced preference graph
and/or the query size become larger, the
time for computing the semantics increases.
However, it is worth noting that semantics
of the query depends on the structure of the
graph and, hence, in certain cases, it may
be possible that the semantic computation in 
a larger graph or for a larger query
takes less time than the computation in
a smaller graph or for a smaller query. For instance,
we observe that computation of query 
$\mlangpref{\true}{(\mlangpref{\true}{true}{\{1,2,3\}})}{\{4,5,6\}}$ takes less time in most
cases than the computation of query
$\mlangpref{\true}{(\mlangpref{\true}{\true}{\{1,2\}})}{\{2,3\}}$ (even if the former involves
$6$ stakeholders). This can be attributed
to situations where the nested query $\mlangpref{\true}{\true}{\{1,2,3\}}$ in 
$\mlangpref{\true}{(\mlangpref{\true}{\true}{\{1,2,3\}\}})}{\{4,5,6\}}$ returns
a small set (or even an empty set), which
makes the evaluation of overall query
computationally less expensive. 

Recall that the result is an average of timing results
obtained from $25$ randomly generated induced
preference graphs for each configuration. It is
worth noting that the maximum time recorded
in all sample runs is $55$ seconds, which corresponds
to a sample for configuration $30,\ 400,\ 400$
for evaluation of query with nesting depth $3$.

\begin{table*}[bth]
\scriptsize
\begin{tabular}{|l||l|l|l|l||l|l|l|l||l|l|l|l|}
\hline
Configuration &
\multicolumn{4}{|c|}{$\mlangpref{\true}{(\mlangpref{\true}{\true}{1,2})}{3,4}$} &
\multicolumn{4}{|c|}{$\mlangpref{\true}{(\mlangpref{\true}{\true}{1,2,3})}{4,5,6}$} &
\multicolumn{4}{|c|}{$\mlangpref{\true}{(\mlangpref{\true}{(\mlangpref{\true}{\true}{1,2,3})}{4,5,6})}{7,8,9}$} \\ \hline 
&
$\texttt{W}_1\texttt{A}_2$ &
$\texttt{W}_1\texttt{A}_1$ &
$\texttt{W}_2\texttt{A}_2$ &
$\texttt{W}_2\texttt{A}_1$ &
$\texttt{W}_1\texttt{A}_2$ &
$\texttt{W}_1\texttt{A}_1$ &
$\texttt{W}_2\texttt{A}_2$ &
$\texttt{W}_2\texttt{A}_1$ &
$\texttt{W}_1\texttt{A}_2$ &
$\texttt{W}_1\texttt{A}_1$ &
$\texttt{W}_2\texttt{A}_2$ &
$\texttt{W}_2\texttt{A}_1$ 
\\ \hline \hline
$10,\ 100,\ 200$ &
0.1589 &
0.1656 &
0.1684 &
0.1753 &

0.1811 &
0.194 &
0.1913 &
0.2049 &

0.7936 &
0.7702 &
0.8762 &
0.8997
\\ \hline

$20,\ 100,\ 200$ &
0.349 &
0.3607 &
0.4912 &
0.5018 &

0.0827 &
0.0873 &
0.0671 &
0.0698 &

1.1415 & 
1.3145 &
1.6448 &
1.8579  \\ \hline 

$30,\ 100, \ 200$ & 
0.6637 &
0.6284 &
0.5505 &
0.5023 &

0.1592 &
0.1807 &
0.1365 & 
0.137 &

1.7729 &
1.9391 &
3.612 &
3.837 \\ \hline \hline \hline

$10,\ 200, \ 200$ &
0.1594  &
0.1667 &
0.1469 &
0.158 &

0.1637 &
0.177 &
0.1433 &
0.1518 &

0.8713  &
0.9955 &
0.8776 &
0.9767
\\ \hline

$20,\ 200, \ 200$ &
0.3364 & 
0.3366 &
0.5669 &
0.5571 & 

0.311 &
0.311 &
0.2698 &
0.2792 &

1.9861 &
1.8095 &
6.674 &
6.2761 \\ \hline

$30,\ 200, \ 200$ & 

0.3097 &
0.306 &
0.2612 &
0.2654 &

0.3357 &
0.3544 &
0.2858 &
0.2913 &

1.5188 &
1.5963 &
1.5699 &
1.4971 

\\  \hline \hline \hline

$20,\ 200, \ 400$ &
0.4341 &
0.4514 &
0.4767 &
0.493 &

0.3863 &
0.4027 &
0.3632 &
0.3709 &

5.53 &
6.0667 &
5.2822 &
5.5943

\\ \hline

$30,\ 200, \ 400$ &
1.4318 &
1.1996 &
3.6899 &
3.4727 &

3.5117 &
3.603 &
4.371 &
4.3949 &

16.0171 &
17.0694 &
20.4699 &
20.7602

\\ \hline

\end{tabular}
\caption{Timing Results}
\label{tbl:timing}
\end{table*}
\section{Summary and Discussion}
\label{sec:conc}
\noindent\textbf{Summary. }We provided the first formal treatment of reasoning with multi-stakeholder preferences in a setting where each stakeholder expresses their preferences in a qualitative preference language. We introduced a query language for expressing queries with respect to the preferences of a given set of stakeholders over sets of outcomes. Motivated by the needs of application scenarios, we introduced and analyzed several alternative semantics for such queries and examined their inter-relationships.  We provided a provably correct algorithm for answering multi-stakeholder preference queries using model checking in alternation-free $\mu$-calculus. Results of preliminary experiments demonstrate the feasibility of the approach.\\

\noindent\textbf{Related Work. }  Existing approaches to reasoning about qualitative preferences of multi-stakeholders leverage voting-based social choice mechanisms~\cite{mcpnet,rossi:synthesis2011,mpcpnet}, starting with the seminal work of Rossi et al. \cite{mcpnet}.  The applicability of such approaches is limited to settings where the stakeholder preferences are expressed over outcomes (rather than attributes of outcomes); or when they are expressed over attributes of an outcome, they are rather simple (e.g., expressible using CP-nets). A major focus of the social choice based approaches to multi-stakeholder preference reasoning is on voting strategies that are resistant to manipulation by some of the stakeholders and guarantee {\em fair} outcomes. 
The key aspects of our work that distinguish from social choice model such as mCP-net \cite{mcpnet} are as follows: We seek to answer queries of the form 
$\mlangpref{\psi_1}{\psi_2}{A}$, i.e., identify outcomes that satisfy $\psi_1$ and are more preferred to outcomes satisfying $\psi_2$, and are not less preferred to any outcome that satisfies $\psi_2$ by the set $A$ of stakeholders, whereas mCP-net queries are about whether one outcome is preferred to another by the given set of stakeholders. The precise conditions for deciding the answer to $\mlangpref{\psi_1}{\psi_2}{A}$ depends on the type of semantics. In the special case where the set of outcomes satisfying $\psi_2$  is a singleton set, then our semantics is similar to Pareto semantics defined in \cite{mcpnet}. This raises the possibility of extending voting-based semantics where the set of outcomes satisfying $\psi_2$  is not a singleton set.  In such as setting, one may use voting to identify an outcome (say $o$) that is preferred by a majority of the stakeholders, and include it in the solution set if it is preferred to one of the outcomes satisfying $\psi_2$ (similar to the witness condition in the paper), and all of them are not preferred to $o$ (similar to the agreement condition in the paper), with the pair-wise outcome preferences decided using a voting mechanism.  \\

\noindent\textbf{Discussion. } 
The framework introduced in this paper is especially useful in applications where it is necessary for multiple stakeholders to be able to express, explore and understand the implications of their preferences in  settings where (i) the individual stakeholder preferences are naturally expressed over attributes of outcomes (as opposed to outcomes themselves), and are sufficiently nuanced to require more expressive preference languages e.g., TCP-nets \cite{Brafman:JAIR06} (which involve tradeoffs between conditional preferences), CI-nets \cite{Bouveret:IJCAI2009} (which can express preferences between sets of objects), or their generalizations \cite{Santhanam:book2016}; and (ii) there is a need for explanations of the role played by the preferences of different stakeholders in determining the outcomes of multi-stakeholder deliberations. One can envision extending this approach to allow individual stakeholders, once they understand the impact of their respective preferences, to minimally revise their preferences to arrive at a consensus that might otherwise have eluded them. \\

\noindent\textbf{Work in progress.} Work in Progress aims to (i) consider organizational structures that further constrain how preferences of multiple stakeholders influence outcomes (e.g., preferences of superiors overriding those of subordinates) (ii) generate targeted explanations of the answers to multi-stakeholder preference queries, (iii) support interactive revision of preferences by stakeholders in the search for consensus or compromise, (iv) further optimize the implementation of the multi-stakeholder preference reasoner, and rigorously assess its scalability as a function of the relevant factors, and (v) apply the resulting tools to support multi-stakeholder decision-making in public policy, healthcare, etc.

\ack This work is supported in part by the National Science Foundation through grants IIS 2225823 and IIS 2225824. 

\typeout{}
\bibliography{references}



\end{document}